\documentclass[11pt,a4paper]{article}
\usepackage[hyphens,spaces]{url}
\usepackage[hyperref]{eacl2021}
\usepackage{times}
\usepackage{latexsym}

\usepackage{microtype}

\aclfinalcopy 

\usepackage[utf8]{inputenc}
\usepackage{graphicx}

\usepackage{listings}
\usepackage{amsmath}
\usepackage{wrapfig}
\usepackage{subfig}
\usepackage{caption}
\usepackage{array}
\usepackage{multirow}
\usepackage{tabularx}
\usepackage{eurosym} 
\usepackage{xspace} 
\usepackage{refcount} 
\usepackage{cleveref}

\newcommand{\fasttext}{FastText\xspace}
\newcommand{\langidpy}{langid.py\xspace}

\title{A reproduction of Apple's bi-directional LSTM models\\ for language identification in short strings}

\author{
    Mads Toftrup\thanks{\;\;Equal contribution}\;\;\textsuperscript{\euro},
    {\bf Søren Asger Sørensen\footnotemark[1]\;\;\textsuperscript{\euro}},
    {\bf Manuel R. Ciosici\textsuperscript{\$}}, 
    \and 
    {\bf Ira Assent\textsuperscript{\euro}} \\ 
  \textsuperscript{\euro} Computer Science Department, Aarhus University \\
  \textsuperscript{\$} Information Sciences Institute, \texttt{manuelc@isi.edu} \\
}

\date{}

\begin{document}
\maketitle
\begin{abstract}
    
    Language Identification is the task of identifying a document's language. For applications like automatic spell checker selection, language identification must use very short strings such as text message fragments. In this work, we reproduce a language identification architecture that Apple briefly sketched in a blog post. We confirm the bi-LSTM model's performance and find that it outperforms current open-source language identifiers. We further find that its language identification mistakes are due to confusion between related languages.

\end{abstract}

\section{Introduction}

Automatic Language Identification is the task of identifying a document's language, an essential task for document classification and machine translation~\cite{ling-etal-2013-microblogs}. General-purpose, open-source Language Identification tools like \emph{\langidpy}~\cite{langidpy} and \fasttext~\cite{Grave} are the \emph{de facto} standards for Language Identification in large documents. 

During the last two decades, text messaging and social media have generated large amounts of short plain-text documents. Language identification on partial and complete short texts presents unique challenges~\cite{jauhiainen2019automatic}. Successful Language Identification can support marketing, political, and socioeconomic analyses on large corpora of short texts such as tweets. Such analyses can, for example, study hate speech towards immigrants and women~\cite{basile-etal-2019-semeval} or seek to understand support groups for smoking cessation~\cite{Prochaska447}. 

On a smartphone, Language Identification on short texts can support several features. Language identification of incoming text messages can help virtual assistants read incoming text messages, which can be an essential tool for minorities such as visually impaired multilingual speakers.

Language identification can also help when typing short texts. Identifying language from the first few characters typed (a very short string) can allow a smartphone to select the correct spelling and grammar checker automatically. Such features motivated a team at Apple to study character-level Language Identification using bi-directional LSTMs~\cite{apple}.

This paper reproduces the architecture presented in an industry blog post~\cite{apple} on Language Identification on extremely short strings (10 characters or less). The blog post briefly sketches the language identification system used by Apple's smartphones and computers. However, due to the use of internal, proprietary corpora, the architecture's performance cannot be compared with the current \emph{de facto} standards for Language Identification: the open-source tools \emph{\langidpy}~\cite{langidpy} and \emph{\fasttext}~\cite{fasttext,joulin2016fasttext,Grave}. 

Our reproduction confirms the performance described in the original blog post \cite{apple}. We go beyond mere reproduction and (1) compare the bi-LSTM model with the current \emph{de facto} standards for Language Identification and (2) analyze performance on related languages.  We find that the bi-LSTM is more accurate than out-of-the-box \fasttext and \langidpy, even outperforming the re-trained \fasttext. Our results suggest that the bi-LSTM architecture could be an alternative to \fasttext and \langidpy for Language Identification on short strings.\footnote{\label{ftnote:source_code}Our source code and models are available at \url{https://github.com/AU-DIS/LSTM_langid}. End-users can download our code as a library from the Python Package Index (PyPI) via \url{https://pypi.org/project/LanguageIdentifier/}.}

\section{Related work}

The simplest Language Identification methods discriminate using elementary distinguishing traits like unique character combinations, frequent or unique words, diacritics, or common n-grams \cite{dunning1994statistical,souter1994natural,TruicaDiacritics}. Increasing model complexity, some Language Identification methods model sequences of words, characters, or bytes. Some methods focus on modeling the frequency of n-grams, e.g., frequency of character n-grams~\cite{applengram,souter1994natural}. Such methods outperform techniques based on unique words. Markov model-based approaches estimate the probability of a string based on n-grams of characters or bytes~\cite{dunning1994statistical}, as is the case of \langidpy~\cite{langidpy,lui-baldwin-2011-cross}. Due to its availability as an open-source library, \langidpy is one of the most popular language identifiers. 

Recent language identifiers increasingly use word representations. For example, in a blog post,  \citet{Grave} shows how to identify languages using \fasttext vectors~\cite{Bojanowski2016a,fasttext,joulin2016fasttext}, which model character n-grams. Language identification with \fasttext vectors is as performant as \langidpy~\cite{Grave}. Similar to \langidpy, \fasttext language identification models are open-source and, therefore, popular. 

LanideNN~\cite{LanideNN} identifies languages in multilingual documents using a recurrent neural network with a single layer of gated recurrent units (GRU). Unlike Markov-based methods, recurrent neural network architectures do not model character sequences with a fixed window of context. 
The language identifier that \citeauthor{apple} briefly sketched in a blog post \cite{apple} uses a recurrent neural network with a two-layer bi-directional LSTM to model character sequences. {\protect\NoHyper\citeauthor{apple}\protect\endNoHyper}'s method differs from LanideNN in architecture complexity (two layers, LSTM cells instead of the simpler GRU cells) and in its focus. LanideNN works with long multilingual documents, whereas \citeauthor{apple} classify extremely short monolingual strings.

In a survey, \citet{jauhiainen2019automatic}  present more than the techniques above, discuss challenges, and identify remaining research questions. Among the remaining research questions are very short texts (the problem motivating \citeauthor{apple}) and discrimination of related languages. In this paper, we go beyond reproducing \citeauthor{apple}'s work by analyzing the effect of related languages.

\section{Model architecture}

\begin{figure}
    \centering
    \includegraphics[width=0.48\textwidth]{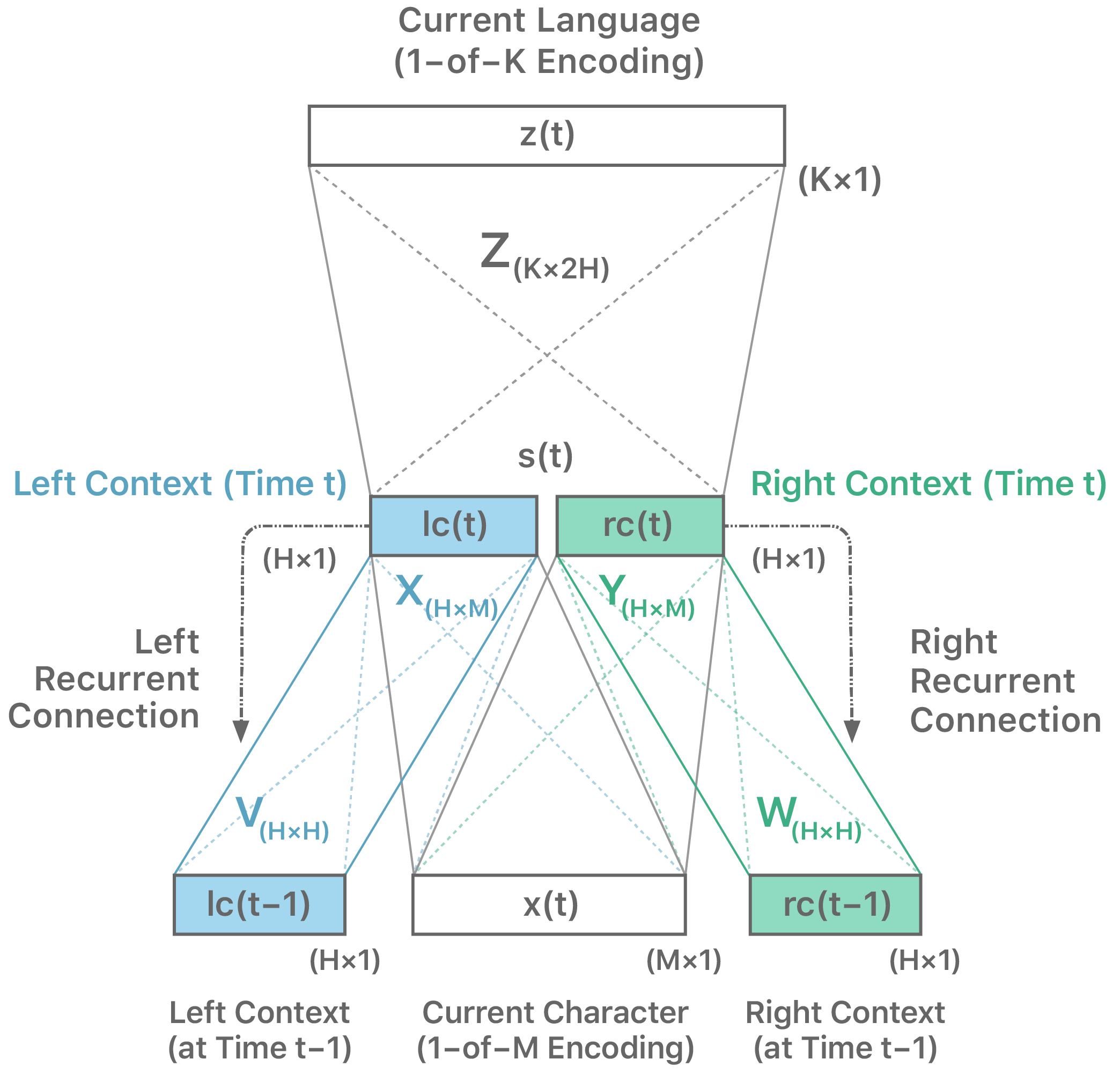}
    \caption{The bi-LSTM architecture. Figure reproduced from \citet{apple}.}
    \label{fig:architecture}
\end{figure}

\Cref{fig:architecture} gives an overview of the two-layer bi-directional LSTM architecture powering Apple's products, as briefly sketched in a blog post \cite{apple}.

The model takes as input strings of characters. In the following, we describe the left-to-right direction of the bi-directional LSTM. The right-to-left direction is identical but mirrored. In the first step, vector embeddings replace all characters in the input string. The network uses a single embedding for all languages since the language is unknown at this point. At each time step, the LSTM ingests a character's embedding and the hidden layer representation from the previous step. The per-character output from the left-to-right LSTM layer is concatenated with that of the right-to-left layer. The concatenated vectors pass to a second LSTM layer that is identical to the first but does not share parameters. After the second layer, the concatenated vectors go through a single linear layer, producing a distribution over all supported languages. The linear layer provides character-level language identification. In other words, for each input character, the network generates a probability distribution over the possible languages.

With the outputs from the linear layer, \citet{apple} state that \emph{A max pooling style majority voting decides the dominant language of the string}. However, max pooling and majority voting are different techniques. A combination of the two is impossible as one cannot perform majority voting over outputs that have been max pooled, and vice versa. Instead, we sum over the linear layer's output values at each time step and \emph{softmax} the summed output to obtain a prediction. We expect this approach to be what the original authors intended. The similarity between our reproduction's performance and what \citeauthor{apple} report in the original blog post confirms our approach.

\section{Data sets}

\Citet{apple} only mention the kind of data used in their experiments. Therefore, we use two large and openly available data sets of the same kind as \citeauthor{apple}: a subset of OpenSubtitles~\cite{lison2016opensubtitles2016} to study performance on dialog; and Universal Dependencies~\cite[UD, ][]{ud25} for prose. Following \citeauthor{apple}, we trim strings to 50 characters per sample, with all samples starting at the beginning of a word, and remove special characters.

\citeauthor{apple} test on 20 languages that use the Latin alphabet, but only show results on 9 of the 20 and do not specify the remaining 11 languages. Besides the 9 languages in the original blog post, we select 11 languages, some of which are closely related. Thus, our experimental setup\footnote{The languages we use are: Catalan (ca), Czech (cs), Danish (da), French (fr), German (de), English (en), Spanish (es), Estonian (et), Finnish (fi), Croatian (hr), Hungarian (hu), Italian (it), Lithuanian (lt), Dutch (nl), Norwegian (no), Portuguese (pt), Polish (pt), Romanian (ro), Swedish (sv), and Turkish (tr).} is similar to \citeauthor{apple}'s. Including closely related languages increases our data sets' difficulty but supports more interesting and more representative experiments. Specifically, it supports performance analysis on related languages, an open research question~\cite{jauhiainen2019automatic}.

\section{Experiments and results}

We use five-fold cross-validation in all experiments. Following \citet{apple}, we evaluate on strings of 10 characters. We test all models on the same strings.

We use the AdamW optimizer with default parameters in PyTorch; we set the character embedding dimension to 150 and the bi-LSTM's hidden dimension to 150; we train for 25 epochs using batches of 64 examples and use weighted cross-entropy for the loss function.

Out-of-the-box, \fasttext and \langidpy can identify more than our set of 20 languages. For fair evaluation, we limit the set of languages that the models output. For \langidpy, we use a built-in method that limits the number of languages under consideration. For \fasttext, we take the probability distribution over all language predictions, extracting only the relevant 20. We use the large pre-trained \fasttext model\footnote{Available at \url{https://fasttext.cc/docs/en/language-identification.html}}. When re-training \fasttext, we use 15 epochs, with a minimum n-gram length of one character and a maximum of six characters. We leave all other parameters at their default.

\subsection{Comparison with original work}

\begin{figure}
    \centering
    \includegraphics[width=0.4\textwidth]{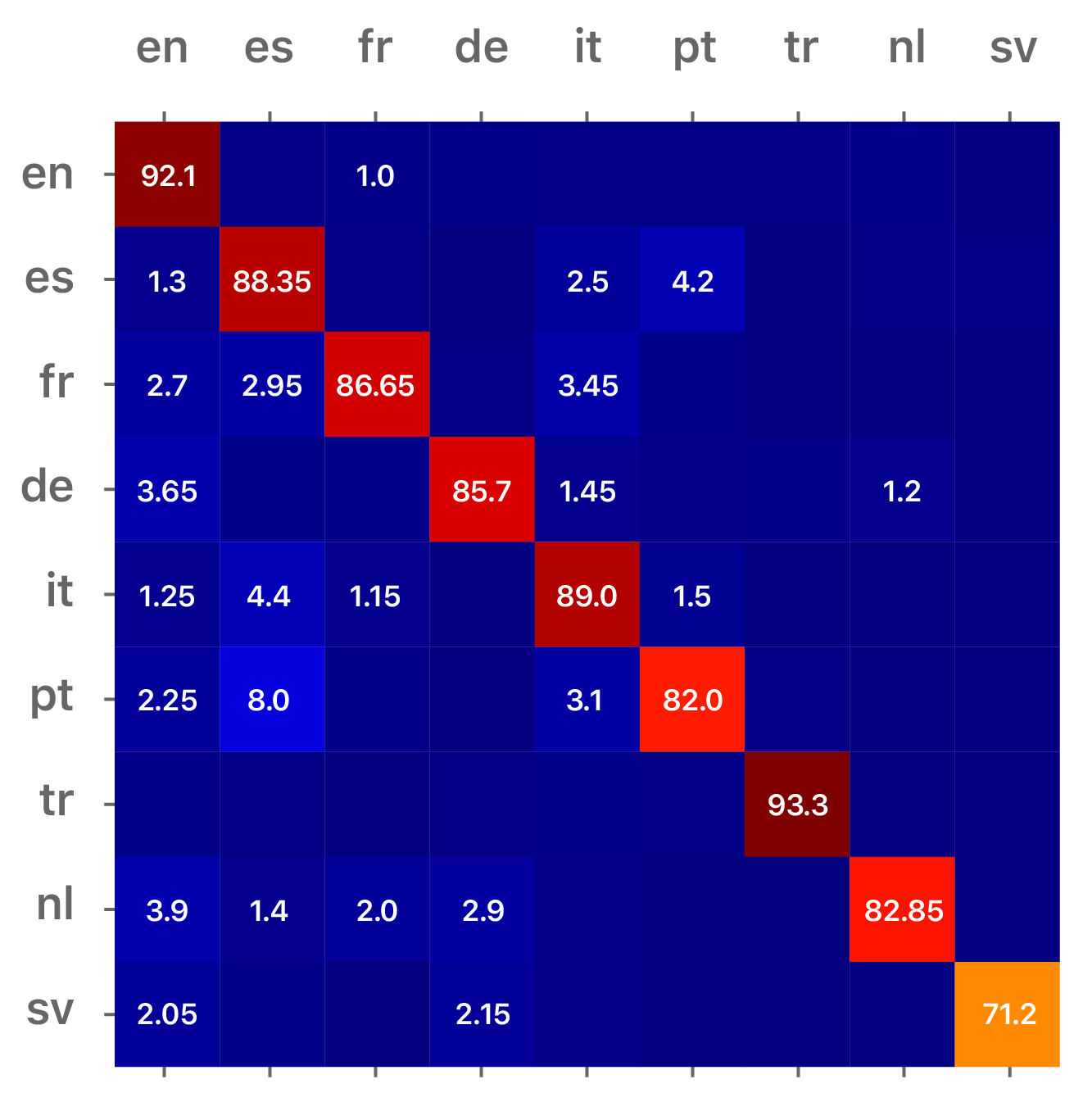}
    \caption{\citet{apple}'s original results.}
    \label{fig:apple_confusion_9_latin}
\end{figure}

\begin{figure*}
    \centering
    \begin{minipage}{.5\textwidth}
        \centering
        \includegraphics[width=0.8\textwidth]{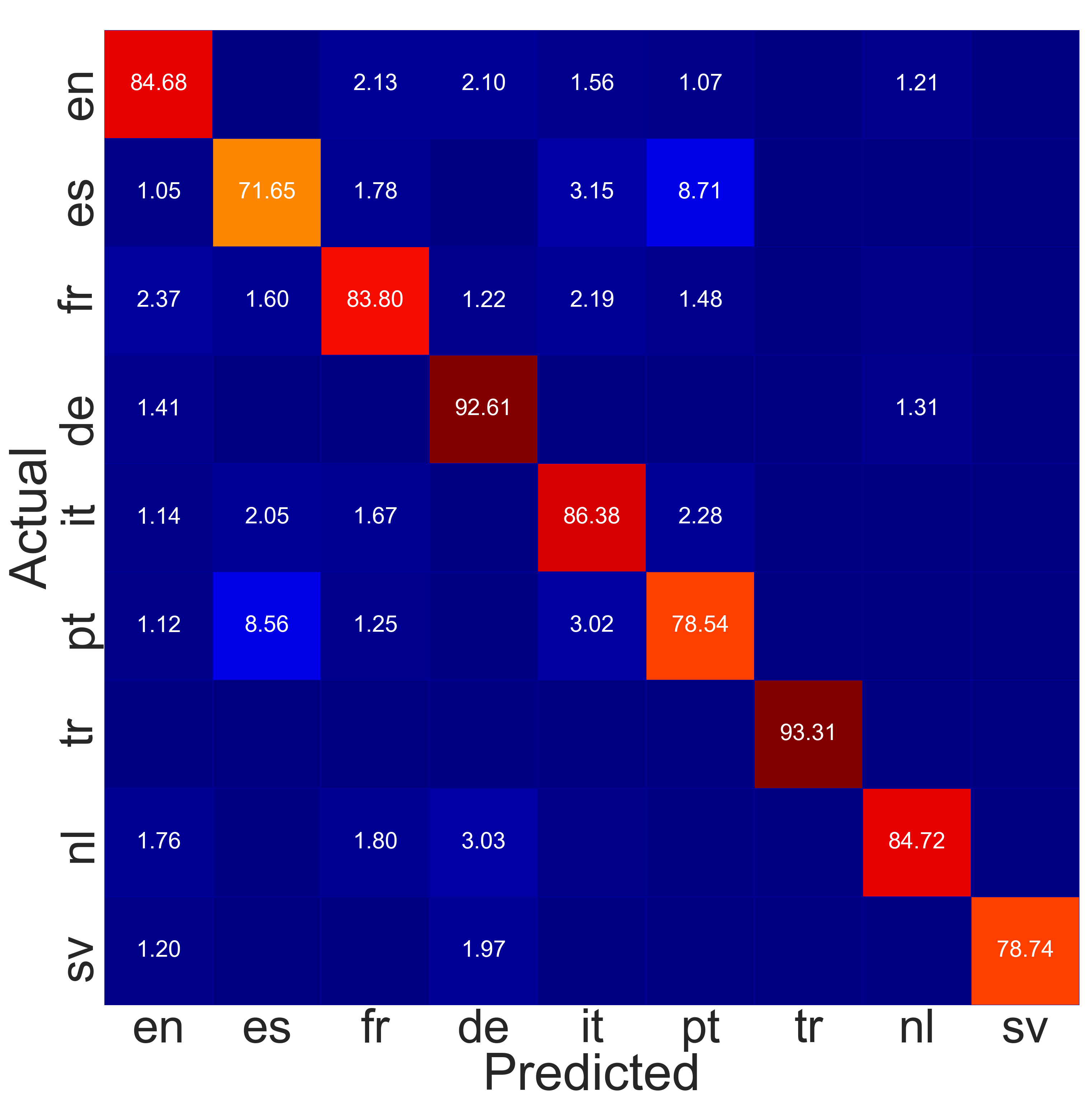}
        \captionof{figure}{Confusion matrix for bi-LSTM on UD.}
        \label{fig:conf_ud9_bilstm}
    \end{minipage}%
    \begin{minipage}{.5\textwidth}
        \centering
        \includegraphics[width=0.8\textwidth]{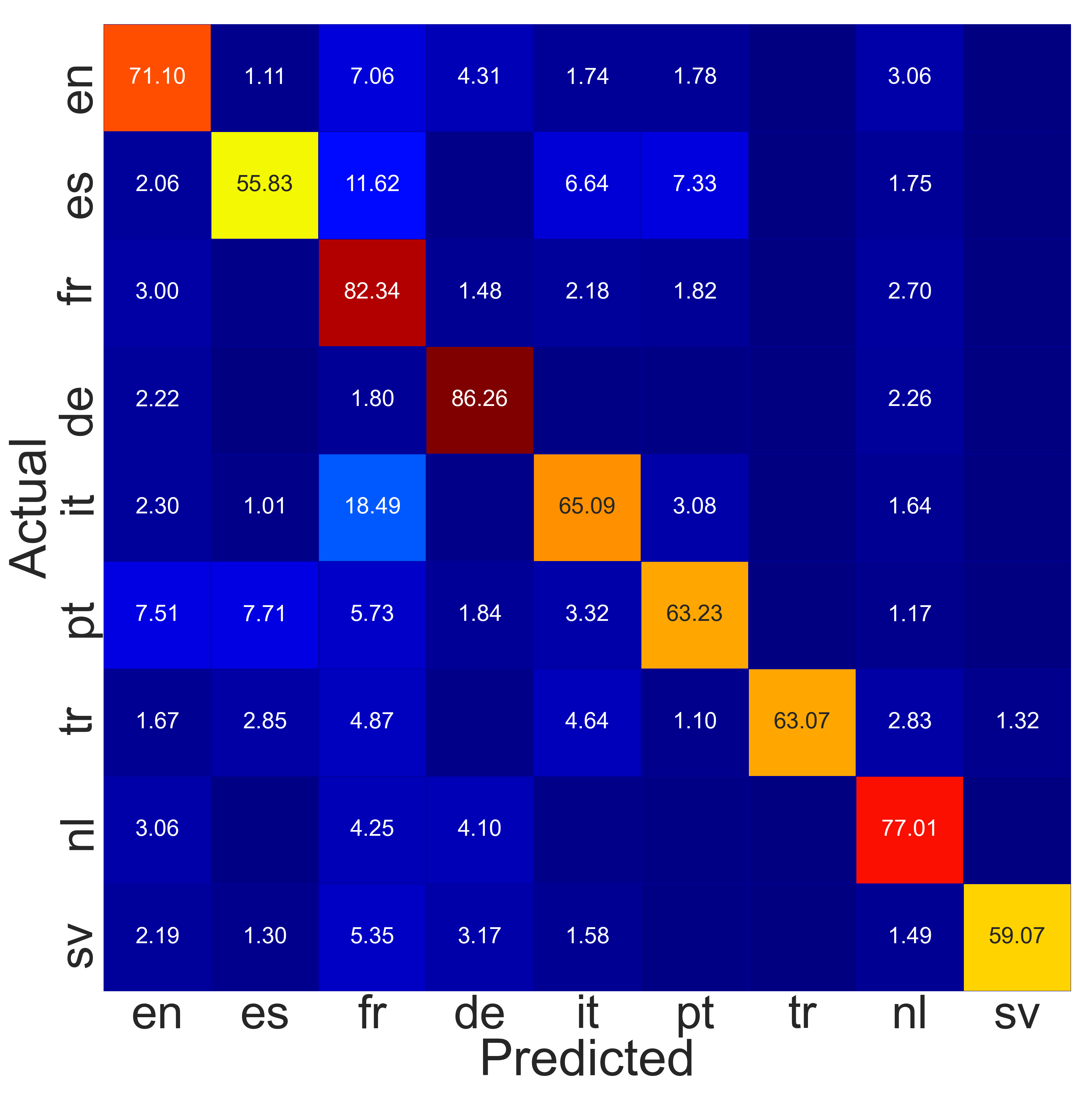}
        \captionof{figure}{Confusion matrix for re-trained \fasttext on UD.}
        \label{fig:conf_ud9_fasttext}
    \end{minipage}
\end{figure*}

\Cref{fig:conf_ud9_bilstm} contains the results of our reproduction of the experiment in Figure (b) from \citet{apple}, a confusion matrix of the bi-LSTM model trained and evaluated on the UD data set. Since \citeauthor{apple} do not include averaged results, we use the confusion matrices for comparison. \Cref{fig:apple_confusion_9_latin} includes a copy of Figure (b) from \citet{apple} for easier comparison. We find that performance per language is similar between the two implementations. While in one case, accuracy is almost identical (Turkish, tr), for most languages, our implementation is either a few points of accuracy below (e.g., French, fr, $-2.85$ points, and Italian, it, $-2.62$) or above the original model (e.g., Dutch, nl, $+1.87$). For some languages, our implementation considerably underperforms the original (e.g., English, en, $-7.4$ points, and Spanish, es, $-16.7$). Our implementation considerably outperforms the original  on German (de $+6.91$) and Swedish (sv $+7.54$). We attribute the difference in performance to randomness during training and differences in training data. The original blog post does not state the size nor language composition of the data set.

In \Cref{fig:conf_ud9_bilstm}, we follow \citeauthor{apple} and threshold values in the confusion matrix at $1.0$. Thus, we can effortlessly compare error patterns. Interestingly, the patterns are almost identical. Both matrices show issues distinguishing between Italian~(it) and Portuguese~(pt), German~(de) and Dutch~(nl), French~(fr) and English~(en), and Italian~(it) or Portuguese~(pt) vs. Spanish~(es) or French~(fr). Unsurprisingly, most confusions appear for languages from the same families, Romance~(es, fr, it, pt) and Germanic~(de, nl).

\subsection{Comparative analysis}

In \Cref{tab:ud20,tab:os20}, we include the comparative analysis results with the current \emph{de facto} standards for Language Identification: \fasttext and \langidpy. We use two weighing strategies for F1 to provide different insights. \mbox{Macro-F1} averages the per-language results and considers languages equally important. \mbox{Weighted-F1} takes into account the popularity of the different languages in the data sets. \mbox{Weighted-F1} measures the performance on the data set, while \mbox{macro-F1} illustrates language coverage as it is not affected by label frequency. In \mbox{multi-class} classification, \mbox{micro-F1} equals accuracy. We, therefore, include only accuracy, denoted \emph{acc@1}.

On both data sets, the bi-LSTM exceeds the weighted- and macro-F1 of \langidpy, pre-trained \fasttext, and re-trained \fasttext. The performance difference between the bi-LSTM and the next best model (the re-trained \fasttext) also appears in the confusion matrix. \Cref{fig:conf_ud9_fasttext} shows that even the re-trained \fasttext exhibits confusion across all pairs. It also shows a strong bias towards some languages like English (en), French (fr), or Dutch (nl) regardless of the input language. All columns in \Cref{fig:conf_ud9_fasttext} that correspond to these languages exhibit confusion errors.

\begin{table}
    \begin{tabular}{|l|l|l|l|l|}
        \hline
        & \textbf{LSTM} & \textbf{pFT} & \textbf{rFT} & \textbf{\langidpy}\\ \hline
        wF1    & \textbf{87.41} & 72.45 & 78.67 & 64.89 \\ \hline
        maF1   & \textbf{79.22} & 61.20 & 67.90 & 51.66 \\ \hline
        acc @1 & \textbf{86.93} & 70.45 & 77.92 & 61.73 \\ \hline
        acc @3 & \textbf{96.07} & 85.84 & 90.59 & 82.83 \\ \hline
        acc @5 & \textbf{97.78} & 90.92 & 94.45 & 88.99 \\ \hline
    \end{tabular}
    \caption{Results on UD. pFT~=~pre-trained \fasttext; rFT~=~re-trained \fasttext}
    \label{tab:ud20}
\end{table}

\begin{table}
    \begin{tabular}{|l|l|l|l|l|}
        \hline
        & \textbf{LSTM} & \textbf{pFT} & \textbf{rFT} & \textbf{\langidpy}\\ \hline
        wF1    & \textbf{91.38} & 67.45 & 84.14 & 54.31 \\ \hline
        maF1   & \textbf{91.38} & 67.45 & 84.14 & 54.31 \\ \hline
        acc @1 & \textbf{91.37} & 67.73 & 84.13 & 53.47 \\ \hline
        acc @3 & \textbf{98.14} & 84.15 & 95.08 & 76.30 \\ \hline
        acc @5 & \textbf{98.93} & 89.31 & 97.38 & 84.22 \\ \hline
    \end{tabular}
    \caption{Results on OpenSubtitles. pFT~=~pre-trained \fasttext; rFT~=~re-trained \fasttext}
    \label{tab:os20}
\end{table}

\begin{figure*}
    \centering
    \includegraphics[width=0.9\textwidth]{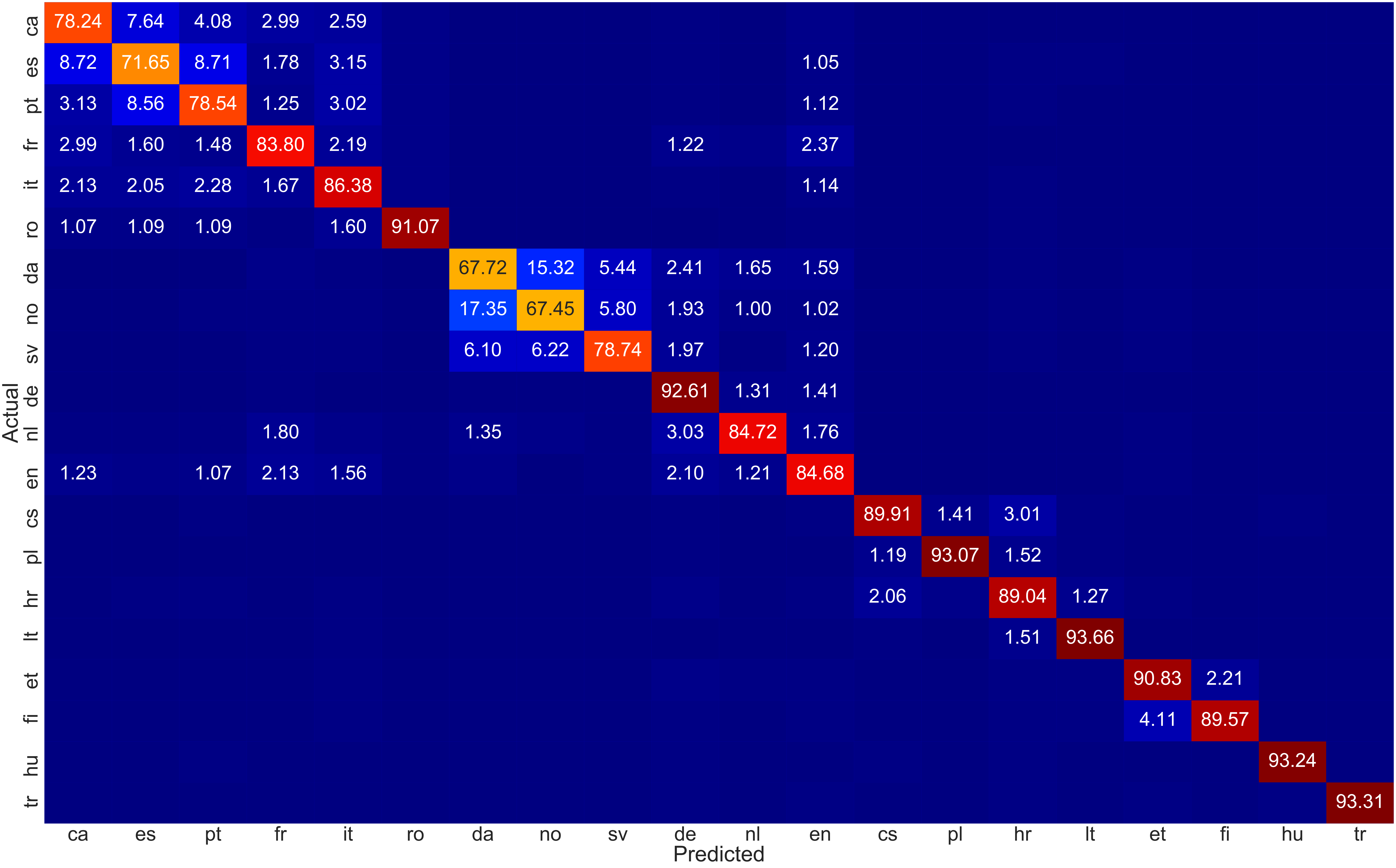}
    \caption{Confusion matrix for bi-LSTM on UD.}
    \label{fig:conf_ud20_bilstm}
\end{figure*}

The OpenSubtitles data is more challenging than UD for out-of-the-box \langidpy and \fasttext, but easier for bi-LSTM and re-trained \fasttext. Also, there is a considerable improvement from the pre-trained \fasttext to the re-trained \fasttext on both data sets. These observations suggest that (1) domain adaptation has a considerable impact on \fasttext, and (2) that dialog is more difficult for the out-of-the-box models. OpenSubtitles contains subtitles of movies predominantly produced in English. Consequently, character names are also English-centered, e.g., Jane. Character names can appear in dialog, which might confuse the pre-trained models to assign such dialog lines to English, despite their translation.

\subsection{Error analysis}

\Cref{tab:ud20,tab:os20} show a jump from accuracy at the top of the list of prioritized predicted languages~(\emph{acc@1}) to accuracy at the top three~(\emph{acc@3}). For most models, a smaller jump follows to accuracy at the top five~(\emph{acc@5}). The sizeable jump indicates that, even when the models are wrong, the correct answer is usually among the top three. For example, from \emph{acc@1} to \emph{acc@3}, the bi-LSTM jumps $9.14$ points on UD and $6.77$ on OpenSubtitles, but only $1.71$ and $0.79$ from \emph{acc@3} to \emph{acc@5}. The gap from \emph{acc@1} to \emph{acc@3} is much larger for \langidpy and \fasttext, illustrating a higher confusion. Recent work in language identification suggests that the accuracy gap might be a symptom of confusion of related languages~\cite{haas2020discriminating}.

To understand the bi-LSTM's jump in accuracy, we turn to the complete confusion matrix. In \Cref{fig:conf_ud20_bilstm}, we show the confusion matrix of the bi-LSTM on all 20 languages in our experiments. There is intense confusion between highly similar languages. We observe three large clusters of confused languages: Romance (ca, es, fr, it, pt, ro), West Germanic (de, en, nl), and  languages of Northern Europe (da, no, sv). More closely related languages are more confusing, for example, Catalan (ca) vs. Spanish (es) and Danish vs. Norwegian (no). The clusters of confusion between related languages indicate that, despite the bi-LSTM's improved performance, highly similar languages still pose a challenge.

 \subsection{Storage requirements}
 \citet{apple} also consider storage requirements. Our bi-LSTM uses $4$~MB of storage, confirming the claims in the original blog post. The re-trained \fasttext model requires $1.5$~GB of storage, but that could reduce to approximately $150$~MB, following \citet{joulin2016fasttext}. \langidpy's model is only $2.5$~MB. Given its language identification performance and model size, the bi-LSTM is a great value proposition, especially on storage-constrained mobile devices, confirming \citeauthor{apple}'s use case scenario.

\section{Conclusions}

We have reproduced the bi-LSTM language identification architecture described in a blog post by \citet{apple}. Our reproduction experiments confirm the performance claims in the original blog post. We evaluated the bi-LSTM against the \emph{de facto} open-source language identifiers in experiments on two openly available data sets. Our evaluation considered dialog and prose, and targeted twenty languages, including some highly similar languages such as Danish (da) and Norwegian (no) or Catalan (ca) and Spanish (es). Our experiments illustrate the diﬃculty of identifying the language in very short strings. The reproduced bi-LSTM outperformed \fasttext and \langidpy on all measures, even when training \fasttext on the same data. However, we went beyond a straightforward reproduction and considered related languages. Our analysis shows that the bi-LSTM can easily confuse languages from the same family~(e.g., Romance, West Germanic, or Scandinavian) and highly similar languages such as Catalan~(ca) and Spanish~(es). We publish our implementation's source code and make a trained model available as a library. In the future, we would like to consider avenues for improving the bi-LSTM architecture. For example, we would like to replace the majority voting mechanism in the bi-LSTM with a more robust alternative.

\bibliography{bibliography}
\bibliographystyle{acl_natbib}

\end{document}